\documentclass[runningheads]{llncs}
\usepackage[T1]{fontenc}
\usepackage{amsmath,amsfonts}
\usepackage{amssymb}
\usepackage{algorithm}
\usepackage{makeidx}
\usepackage{graphicx}
\usepackage{epstopdf}
\usepackage{diagbox}
\usepackage{booktabs}
\usepackage[usenames,dvipsnames]{color}
\usepackage{subcaption}
\usepackage{caption}
\usepackage{multirow}
\usepackage{url}

\usepackage{graphicx}
\begin{document}
\title{Topological Cycle Graph Attention Network for Brain Functional Connectivity}
\titlerunning{CycGAT}
\author{Anonymous}
\authorrunning{Anonymous}
\institute{Anonymous Organization\\
\email{**@*****.***} \\
}

\author{Jinghan Huang\inst{1} \and
Nanguang Chen\inst{1} \and
Anqi Qiu\inst{1-5}}
\authorrunning{J. Huang et al.}

\institute{$^{1}$ Department of Biomedical Engineering, National University of Singapore, Singapore\\
$^{2}$  Department of Health Technology and Informatics, The Hong Kong Polytechnic University, Hong Kong\\
\email{an-qi.qiu@polyu.edu.hk} \\
$^{3}$ NUS (Suzhou) Research Institute,  National University of Singapore,  China \\
$^{4}$  The N.1 Institute for Health, National University of Singapore, Singapore \\
$^{5}$ Department of Biomedical Engineering, The Johns Hopkins University,  USA 
}
\maketitle              
\begin{abstract}
This study, we introduce a novel Topological Cycle Graph Attention Network (CycGAT), designed to delineate a functional backbone within brain functional graphs—key pathways essential for signal transmission—from non-essential, redundant connections that form cycles around this core structure. We first introduce a cycle incidence matrix that establishes an independent cycle basis within a graph, mapping its relationship with edges. We propose a cycle graph convolution that leverages a cycle adjacency matrix, derived from the cycle incidence matrix, to specifically filter edge signals in a domain of cycles. Additionally, we strengthen the representation power of the cycle graph convolution by adding an attention mechanism, which is further augmented by the introduction of edge positional encodings in cycles, to enhance the topological awareness of CycGAT. We demonstrate CycGAT's localization through simulation and its efficacy on an ABCD study's fMRI data (n=8765), comparing it with baseline models. CycGAT outperforms these models, identifying a functional backbone with significantly fewer cycles, crucial for understanding neural circuits related to general intelligence. Our code will be released once accepted.

\end{abstract}
%
%
%
\newcommand{\cycle}[1]{\mathcal{C}_{#1}}
\def\mygraph{\mathcal{G}}
\def\ecproj{\mathbf{T}}
\def\edgenormal{\mathbf{H}_\mathcal{E}}
\def\cyclenormal{\mathbf{H}_\mathcal{C}}
\def\enecproj{\ecproj\edgenormal^{-1}} 
\def\cnecproj{\cyclenormal^{-1}\ecproj} 
\def\adje{\mathbf{A}_\mathcal{E}}
\def\adjc{\mathbf{A}_\mathcal{C}}
\def\lapc{\boldsymbol{\mathcal{L}}_\mathcal{C}}
\def\relu{\operatorname{ReLU}}
\def\lrelu{\operatorname{LeakyReLU}}
\def\posE{\mathbf{P}_\mathcal{E}}
\def\posC{\mathbf{P}_\mathcal{C}}
\def\firstHL{\mathbf{\mathcal{L}}_1}
\def\EdgeSig{\mathbf{X}_\mathcal{E}}
\newcommand{\edgesig}[1]{\mathbf{x}_{e_{#1}}}
\newcommand{\edgepos}[1]{\mathbf{p}_{e_{#1}}}

\section{Introduction}
Functional magnetic resonance imaging (fMRI) is a non-invasive technique capturing brain activity through blood oxygen level dependent (BOLD) signals \cite{glover2011overview}. It evaluates functional connectivity (FC) by computing Pearson correlation of time series between brain regions, offering insights into a brain's functional organization.  FC is commonly modeled as an undirected graph, known as a brain functional graph \cite{fc_Review,xia2023multi}, where nodes represent brain regions and edges denote functional connections. Contrary to being random, these graphs exhibit small-world architectures and significant modularities \cite{suarez2020linking,meunier2009hierarchical}, suggesting the presence of a functional backbone essential for the majority of signal transmission within the brain \cite{sc_backbone}, with other edges considered redundant, forming cycles around this core structure \cite{ryu2023persistent}. The analysis on this backbone contributes to a wider examination of brain connectivity. For example, studies have demonstrated that an increase in signal dispersion across redundant connections correlates with lower cognitive scores in subjects, highlighting the critical role of the functional backbone in cognitive performance \cite{ryu2023persistent}.  Despite these advances in understanding FC, there remains a significant gap in leveraging deep learning to distinguish the functional backbone from redundant connections. This requires a technique to filter the FC based on the unique topological structures formed by cycles.

The use of graph neural networks (GNNs) to learn the features of FC relevant to cognition or mental disorders has seen significant growth \cite{GNN_review,HLHGCNN,huang2022spatio}. GNNs analyze FC as signals on nodes by aggregating connectivities into a vector \cite{dgcn,braingnn,gat}. Innovations in GNN applications, such as attention mechanisms \cite{gat}, clustering-based embeddings \cite{braingnn}, and dynamic network updates \cite{dgcn}, have propelled forward the classification of disorders like Autism Spectrum Disorder (ASD) and Attention Deficit Hyperactivity Disorder (ADHD) \cite{braingnn,dgcn}. Recent advancements have also explored edge signal smoothing through topological edge connections \cite{kawahara2017brainnetcnn,ehgnn,HLHGCNN}, employing strategies like dual graph conversions \cite{ehgnn} and spectral graph convolution with Hodge Laplacian operators \cite{HLHGCNN} to address the challenges of dimensionality and spatial localization. These edge-focused GNN approaches have shown promise in both molecular science and neuroscience.  However, there remains a critical gap in modeling higher-order interactions, such as cycles, which are essential for distinguishing between the functional backbone and redundant connections.


This study introduces the cycle graph attention network (CycGAT), an innovative approach that incorporates the topological concept of cycles into GNNs to filter out redundant connections and extract the functional backbone. CycGAT is designed to refine the analysis of fMRI data for classifying general intelligence groups by learning from the edges' features and their neighborhood connections in a domain of cycles. The effectiveness of CycGAT is highlighted through its application on the large-scale ABCD dataset, where its performance is benchmarked against leading GNN methods such as GAT, BrainGNN, dGCN, Hypergraph NN, and HL-HGCNN. This research introduces three novel techniques:

\begin{enumerate}
\item An efficient algorithm computing a cycle incidence matrix and a cycle adjacency matrix characterizing the formation of cycles by edges and the intricate connections among edges within these cycles. respectively;
\item An attention-based spatial graph convolution operator that smooths edge signals in a domain of cycles;
\item Topological-aware edge positional encodings that represent the topological distance between edges in cycles.
\end{enumerate}

\begin{figure}[htbp]
\begin{center}
\includegraphics[width=1.0\textwidth]{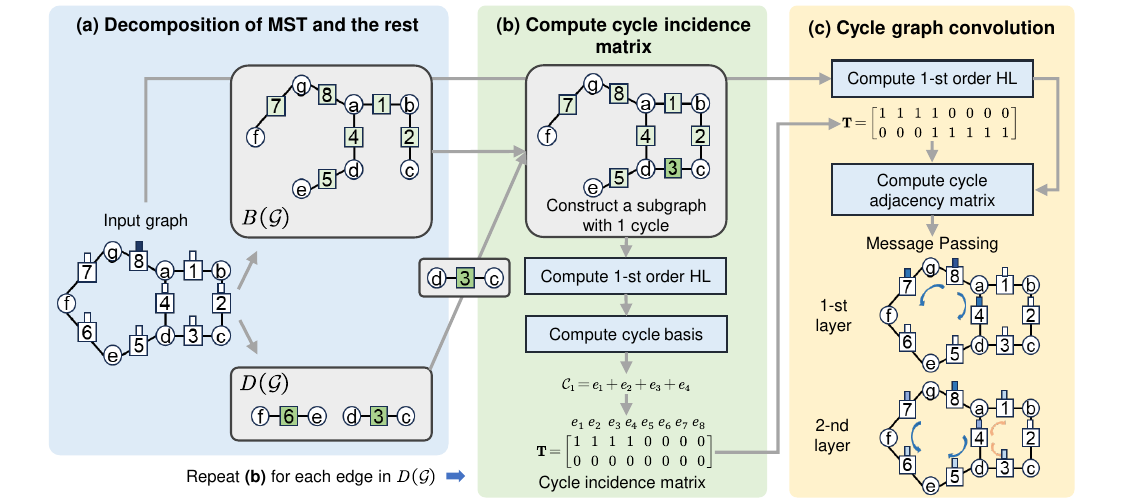}
\caption{CycGAT framework overview. (a) The input graph displays nodes (circles), edges (squares), and edge signals (rectangles), with edge decomposition into a maximum-spanning tree, $B(\mygraph)$, and the set of additional edges, $D(\mygraph)$. (b) Detailing the computation of the cycle incidence matrix, iteratively applied to each edge in $D(\mygraph)$. (c) Depiction of cycle graph convolutional layers, with blue arrows indicating signal propagation within a single cycle and orange arrows demonstrating inter-cycle signal propagation via shared edges.
}  
\label{fig:framework} 
\end{center}
\end{figure} 
\section{Methods}

This study introduces the Cycle Graph Attention Network (CycGAT), a novel framework designed for the automatic identification of functional backbones within brain functional graphs.  In the following,  we will detail the computation processes for the cycle incidence matrix, the cycle adjacency matrix, and the edge positional encodings in cycles (EPEC) in sequence. Finally, we present our newly proposed cycle graph convolution operator.



\subsection{Cycle Incidence Matrix and Cycle Adjacency Matrix}
In this study, the brain functional network is characterized by an undirected graph, $\mygraph=(V, \mathcal{E})$, where brain regions are represented as nodes, $V=\{v_i\}_{i=1}^N$, and their connections as edges, $\mathcal{E}=\{e_{i}\}_{i=1}^E$. Here, $N$ and $E$ represent the number of nodes and edges. Functional connectivity, defined as the Pearson correlation between fMRI time-series of two nodes, is represented as a signal on the connecting edge. Cycles within $\mygraph$ are defined as sets of edges that connect sequences of nodes in a loop. The cycle incidence matrix $\ecproj$ is defined as
\begin{equation}
    [\ecproj]_{qi}= 
\begin{cases}
    \hspace{.8em}1, & \text{if} \;\text{edge $e_i$ is incident with cycle $\mathcal{C}_q$};\\
    \hspace{.8em}0, & \text{otherwise},
\end{cases}
\end{equation}
where $[\ecproj]_{qi}$ denotes the element of $\ecproj$ in the $q$-th row and $i$-th column. 
$\ecproj$ characterizes how edges in $\mygraph$ participate in forming an independent cycle basis.  The $q$-th cycle basis $\cycle{q}$ can be expressed as $\cycle{q} = \sum_{i=1}^E [\ecproj]_{qi} e_i$, where any cycle in $\mygraph$ can be represented as a linear combination $\sum_{q=1}^Q a_q \cycle{q}$. For example, the cycle connecting all nodes in Fig.~\ref{fig:framework}(a) can be expressed as $\mathcal{C}_1-\mathcal{C}_2$.
In addition, $\ecproj$ serves as the foundation for defining the cycle adjacency matrix. This adjacency matrix, in turn, establishes a domain of cycles crucial for the filtering of edge signals.

To compute $\ecproj$, we first identify an independent cycle basis by dividing the edge set $\mathcal{E}$ into two subsets: $B(\mygraph)$, containing edges that form the graph's maximum spanning tree, and $D(\mygraph)$, including the remaining edges (Fig.~\ref{fig:framework}(a)). 
The number of independent cycles $Q$ equals the number of edges in $D(\mygraph)$. Adding an edge $e_k \in D(\mygraph)$ to $B(\mygraph)$ forms a subgraph $\mygraph' = (V, B(\mygraph) \cup \{e_k\})$ containing exactly one cycle (Fig.~\ref{fig:framework}(b)). This cycle can be easily found by computing the eigenvector corresponding to the zero eigenvalues of the first-order Hodge Laplacian $\firstHL$ of $\mygraph'$. The value $1$ is assigned to particular edges in that cycle, as is shown in Fig.~\ref{fig:framework}(b). 

The cycle adjacency matrix $\adje$, a key element in understanding connections within the brain functional network, is computed by identifying edges connected by nodes and ensuring these edges also share the same cycle, as described by the equation:
\begin{equation}
[\adje]_{ij}= 
\begin{cases}
    \hspace{.8em}1, & \text{if} \; [\firstHL]_{ij} \neq 0 \, \text{and} \; [\ecproj^\top \ecproj]_{ij} \neq 0;\\
    \hspace{.8em}0, & \text{otherwise}.
\end{cases}
\end{equation}
$\top$ denotes a matrix transpose. 

\subsection{Edge Positional Encodings in Cycles (EPEC)}
We encode positions of edges in cycles with EPEC to represent the distance between edges through cycles.  We adopt the concept of Laplacian Eigen-positional encodings, prevalent in graph transformers \cite{gps,HLPE,kreuzer2021rethinking}. Briefly, we calculate the Laplacian eigenvectors for each cycle on the cycle basis, followed by projecting these cycle positional encodings onto edges using the proposed cycle incidence matrix. 
An adjacency matrix of cycles is computed as 
$\big[\adjc \big]_{ij}= 
\begin{cases}
    \hspace{.1em}\big[\ecproj \ecproj^{\top} \big]_{ij}\,, & \text{if} \; i \neq j;\\
    \hspace{.1em}0\,, & \text{otherwise}\,.
\end{cases}$
The degree matrix of $\adjc$, a diagonal matrix, is computed as $\big[\mathbf{D}_\mathcal{C}\big]_{ii} = \sum_{q=1}^Q \big[\adjc\big]_{iq}$. Subsequently, the cycle Laplacian operator, $\lapc = \mathbf{D}_\mathcal{C} - \adjc$, is calculated. The positive-semidefinite nature of $\lapc$ allows for its eigendecomposition, from which we select the $K$ smallest non-trivial eigenvectors as cycle positional encodings, denoted $\posC$. We also normalize the cycle incidence matrix $\ecproj$ against the number of cycles each edge is part of, resulting in $\edgenormal$—a diagonal matrix defined by $\big[\edgenormal\big]_{ii} = \big[\ecproj^{\top} \ecproj \big]_{ii}$. Consequently, the EPEC $\posE$ is calculated as:
\begin{equation}
    \posE = (\enecproj)^{\top} \posC,
\end{equation}

\begin{figure}[htbp]
\begin{center}
\includegraphics[width=1.0\textwidth]{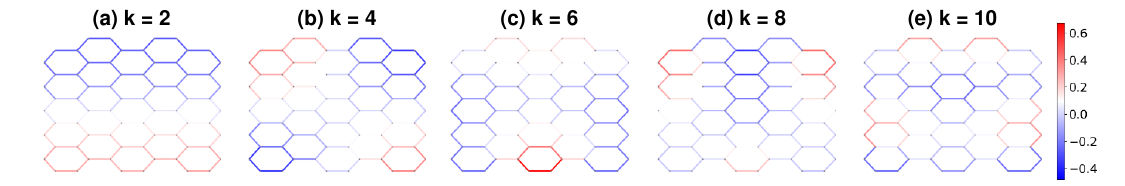}
\caption{Examples of the edge positional encodings in cycles (EPEC) representing the $k$-th order of eigenfunction. The frequencies $\lambda = [0.0148, 0.0296, 0.0581, 0.0799, 0.0977]$, corresponding to panels (a-e), illustrate the EPEC's ability to encode different spatial frequencies. Lower frequencies highlight global and gradual variations within the graph (e.g., top-down orientations), whereas higher frequencies reveal increasingly intricate and localized cycle patterns.
}  
\label{fig:cepe} 
\end{center}
\end{figure}

We use an example to show how EPEC represent the relative positions of edges through cycles (Fig.~\ref{fig:cepe}). The spatial distance between any two edges is determined by the minimum number of cycles they are apart, introducing a natural coordinate system for graphs that mirrors increasing topological complexity.

\subsection{Cycle Graph Convolution}

Cycle graph convolution is formulated by infusing positional encodings and the attention mechanism to the spatial graph convolution in a domain of cycles. Each edge $e_i$ is characterized by its feature $\edgesig{i}\in \mathbb{R}^{d\times 1}$ and EPEC $\edgepos{i}\in \mathbb{R}^{K\times 1}$. The convolution operation on edge $e_i$ is thus defined as:
\begin{equation}
    \edgesig{i}' = \eta\big( \sum_{e_j\in \mathcal{N}_i} \alpha_{ij} \mathbf{W}\edgesig{j} \big),
    \label{eq:MultiHeadConv}
\end{equation}
where $\eta$ signifies a nonlinear activation function, $\mathbf{W} \in \mathbb{R}^{d'\times d}$ is the weight matrix characterizing the spatial filters, and $\mathcal{N}_i$ encompasses the neighboring edges of $e_i$ and $e_i$ itself within $\adje$. The attention coefficient $\alpha_{ij}$, dictating the attention between edges $e_i$ and $e_j$, is computed as:
\begin{equation}
    \alpha_{ij}=\frac
    {\exp{\Big(
     \eta\Big(\mathbf{h}^{\top} \big( \mathbf{W}_1\edgesig{i}  \mathbin\Vert \mathbf{W}_2\edgesig{j}  \mathbin\Vert \mathbf{W}_3 (\edgepos{i}-\edgepos{j} ) \big) \Big) \Big) }}
    {\sum_{e_k\in \mathcal{N}_i} \exp{\Big(
     \eta\Big(\mathbf{h}^{\top} \big( \mathbf{W}_1\edgesig{i}  \mathbin\Vert \mathbf{W}_2\edgesig{k}  \mathbin\Vert \mathbf{W}_3 (\edgepos{i}-\edgepos{k} ) \big) \Big) \Big) }},
\end{equation}
employing the concatenation operator $\mathbin\Vert$. Here, $\mathbf{W}_1$, $\mathbf{W}_2$, $\mathbf{W}_3$, and the vector $\mathbf{h}\in \mathbb{R}^{3d'\times 1}$ are trainable parameters. To further enhance the convolution's capability, a multi-head attention scheme is applied \cite{gat}.

\subsection{Cycle Graph Attention Network (CycGAT)}

The CycGAT architecture, designed with cycle graph convolutional layers, aims to elucidate edge features within FC. Each layer integrates convolution, an activation function, and batch normalization to optimize learning. The leaky rectified linear unit (LeakyReLU) is selected as the activation function, $\eta$, accommodating the biological relevance of negative FC. A final graph convolutional layer applies a sigmoid function as the activation function, transforming edge features into scalar values. This prepares a vectorized edge representation for the output layer, which consists of fully-connected layers to enable classification.

\subsection{Implementation}
\label{sec:Implementation}
The construction of the brain functional graph commences with transforming the connectivity matrix with top 25\% absolute values into a binary matrix. The computation of the Hodge Laplacian follows Huang et al. \cite{HLHGCNN}.  The framework is implemented using Python 3.9.13, Pytorch 1.12.1, and the PyTorch Geometric 2.1.0 library, with the CycGAT model comprising 8 cycle graph convolutional layers, each with 16 filters and 4 heads. Optimization employs the ADAM optimizer, leveraging an NVIDIA Tesla V100SXM2 GPU. Model parameters were optimized using a greedy search approach, with binary cross-entropy as the loss function, L1 penalization on the functional backbone, and an early stopping mechanism based on validation set performance.

This study leverages resting-state fMRI (rs-fMRI) images from the Adolescent Brain Cognitive Development (ABCD) study, which includes 8765 youth aged 9-11 years (\protect\url{https://abcdstudy.org/}). Utilizing the dataset and fMRI preprocessing pipeline as described in Huang et al. \cite{huang2022spatio}, we define nodes as one of 268 brain regions of interest (ROIs) \cite{shen2017using}.
We select subjects with the top and bottom 25\% of General intelligence scores as the high and low general intelligence groups, respectively.



\section{Results}

This section first demonstrates the spatial localization property of the cycle graph convolution in relation to the number of layers via simulated data.  Afterwards, we demonstrate the use of cycle graph convolution and its use in GNN for classifying high or low fluid intelligence groups using the ABCD dataset.  Finally, we visualize the reduction of the number of cycles after cycle graph convolution and the functional backbone.

\subsection{Spatial Localization of the Cycle Graph Convolution}
\begin{figure}[htbp]
\begin{center}
\includegraphics[width=1.0\textwidth]{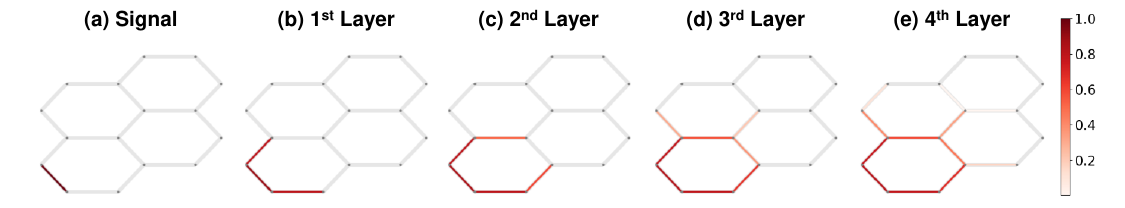}
\caption{CycGAT's spatial localization with four convolutional layers. (a) depicts a pulse signal at a single edge. (b-c) illustrate the signal's propagation within a cycle. (d-e) demonstrate inter-cycle signal propagation through shared edges.}  
\label{fig:localization} 
\end{center}
\end{figure}

We demonstrate the cycle graph convolution's spatial localization by initiating a pulse signal at one edge (Fig.~\ref{fig:localization}(a)) and observing its diffusion through cycle graph convolutional layers. Fig.~\ref{fig:localization}(b-c) illustrate that the signal propagation is confined within individual cycles, highlighting the convolution's spatial localization capability. Crucially, Fig.~\ref{fig:localization}(d-e) reveal how cycles interact through their shared edges, enabling the signal to traverse across different cycles. This mechanism underscores the potential for modeling complex signal interactions and propagation patterns within the brain's functional graph structure.

\subsection{Comparisons with existing GNN methods }
In this section, we compare our model with existing state-of-the-art methods as mentioned above, focusing on the classification accuracy of high and low general intelligence groups using the ABCD dataset. We adopted the architectures of BrainGNN, dGCN, and HL-HGCNN from the corresponding papers \cite{braingnn,dgcn,HLHGCNN}, as all methods are applied to fMRI data. The GAT model is designed with two graph convolutional layers, each consisting of 32 filters and a 2-head attention mechanism, optimized through a greedy search. The input features are the FC vectors of each region. The Hypergraph NN includes two graph convolutional layers with 32 filters and one hypercluster layer added after the first convolutional layer. Our results, as detailed in Table~\ref{comparison_table}, suggest that CycGAT achieves higher accuracy and outperforms all baseline models with all p-values $< 0.05$. To further highlight the performance enhancement provided by the proposed EPEC, we removed the EPEC-related term from Eq.\ref{eq:MultiHeadConv} and constructed a CycGAT variant without EPEC, maintaining the same layer configuration. Table~\ref{comparison_table} indicates that CycGAT with EPEC significantly outperforms its counterpart without EPEC (p=0.005).

\begin{table}[htbp]
\centering
\caption{General intelligence classification accuracy.  $p$-value is obtained from two-sample $t$-tests examining the performance of each method in reference to the proposed CycGAT.
}
\scalebox{0.8}{
\begin{tabular}{p{14em}<{\centering}| p{8em}<{\centering}|p{8em}<{\centering}}
\toprule[1pt]
\textbf{GNN model} & \textbf{Accuracy} & \textbf{$p$-value} \\
\hline 
\textbf{GAT} \cite{gat} & $0.651\pm 0.011$ & $0.001$ \\ 
\textbf{BrainGNN}\cite{braingnn} & $0.656\pm 0.015$ & $0.003$ \\ 
\textbf{dGCN}\cite{dgcn} & $0.646\pm 0.009$ & $0.001$ \\ 
\hline
\textbf{Hypergraph NN}\cite{ehgnn} & $0.665\pm0.014$ & $0.009$ \\ 
\textbf{HL-HGCNN}\cite{HLHGCNN} & $0.674\pm0.011$ & $0.032$ \\ 
\hline
\textbf{CycGAT (without EPEC)} & $0.661\pm0.007$ & $0.005$ \\
\textbf{CycGAT (ours)} & \textbf{0.682$\pm$0.006} &  - \\ 
\bottomrule[1.5pt]
\end{tabular}}
\label{comparison_table}
\end{table}

\subsection{Understanding of the Functional Backbone}
\begin{figure}[htbp]
\begin{center}
\includegraphics[width=1.0\textwidth]{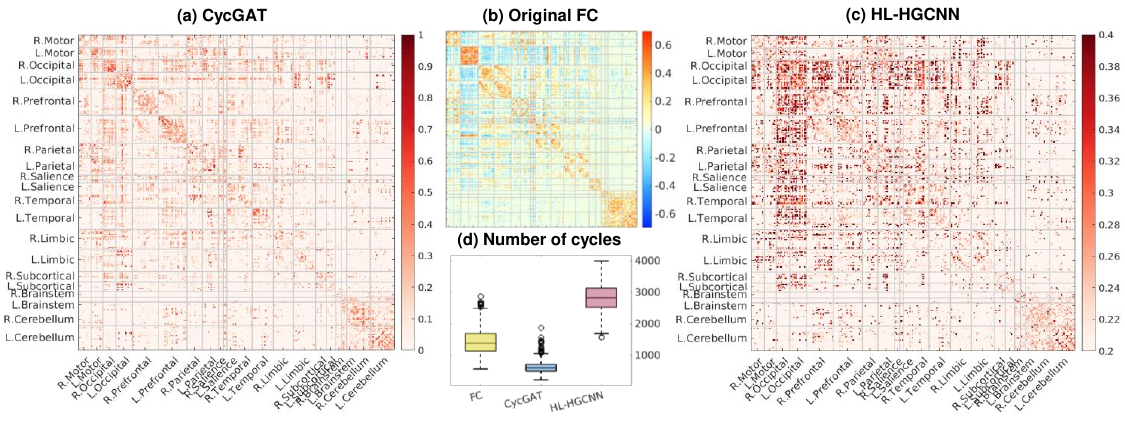}
\caption{Panel (a) and panel (c) show the saliency maps from CycGAT and HL-HGCNN, respectively. Panel (b) shows the brain functional connectivity. The box plot in panel (d) illustrates the number of cycles in FC and saliency maps.
}  
\label{fig:backbone} 
\end{center}
\end{figure}

We visualize the saliency map, which is the output of the final cycle convolutional layer averaging across the test set, as depicted in Fig.\ref{fig:backbone}(a). The saliency map from CycGAT, when compared to the original FC and the saliency map from HL-HGCNN, illustrated in Fig.\ref{fig:backbone}(b-c), exhibits increased sparsity, particularly retaining the functional connectivities within the occipital regions, prefrontal regions, and cerebellum. This observation aligns with established research on general intelligence \cite{GI_occipital,GI_cerebellum,song_iq-2008}.

Furthermore, we quantitatively assess the number of cycles in the saliency map versus the original FC at the subject level, with the quantification method detailed in the supplementary materials. The box plot in Fig.~\ref{fig:backbone}(d) illustrates that CycGAT's saliency map significantly reduces cycles compared to the original FC and HL-HGCNN, with $p<1\times10^{-3}$. HL-HGCNN's use of isotropic filters leads to an increase in cycles due to their over-smoothing nature. Since cycles often imply redundant connections, fewer cycles indicate improved network efficiency, validating our method's effectiveness in identifying the biologically meaningful functional backbone.


\section{Conclusion}
This study proposes a novel CycGAT on functional connectivity for classifying cognitive ability groups.  Our experiments demonstrate the spatial localization property of the cycle graph convolution operator.  In addition, we illustrate how edge positional encodings in cycles serves as edge coordinates that represent the topological relationship between edges.  Moreover,  our CycGAT performs better than the existing state-of-art methods for classifying high and low general intelligence groups,  indicating the potential of our method for future prediction and diagnosis based on fMRI.  Nevertheless,  more experiments on different datasets are needed to further validate the robustness of the proposed model. CycGAT enables further analysis by comparing the functional backbone with structural connectivity, offering insights into structural-functional coupling \cite{suarez2020linking}.

\begin{credits}

\end{credits}

\bibliographystyle{splncs04}
\bibliography{reference.bib}

\end{document}